# Developing and Analyzing Boundary Detection Operators Using Probabilistic Models

David Sher

Department of Computer Science
University of Rochester
Rochester, NY 14627

## 1. Summary

### 1.1. Definition of a Bayesian Feature Detector

Most feature detectors such as edge detectors or circle finders are statistical, in the sense that they decide at each point in an image about the presence of a feature. This paper describes the use of Bayesian feature detectors. A *Bayesian feature detector* is an operator that takes an image, $O$ and returns

(1) The probability that an image identical to the one observed would be seen given that a feature proposition is true $(P(O|F))$.

(2) The probability that the image would be observed independently of the feature proposition $(P(O))$.

An example of a feature proposition is "An object boundary exists at point x,y." Another example of a feature proposition is "Point x,y is in the interior of an object."

Given such an operator and some *apriori* probability for the feature $(P(F))$ the probability that the feature is present given that the image is observed is

$$\frac{P(O|F)P(F)}{P(O)} \quad (1)$$

(by Bayes law). Thus a Bayesian feature detector can be used to calculate the probability of a feature.

$P(O)$ can be calculated as:

$$\sum_{F_i} P(O|F_i)$$

where the $F_i$ are a set of mutually exclusive and all encompassing feature propositions. A set of propositions, $F_i$, are *mutually exclusive* when:

$$\forall i,j (i \neq j \rightarrow P(F_i \& F_j)=0) \quad (2)$$

A set of propositions, $F_i$, are *all encompassing* when:

$$\sum_{F_i} P(F_i) = 1 \quad (3)$$

$F, not\ F$ is always such a set. Bayes law (equation (4)) defines a function that takes the $P(F_i)$ (the priors) to $P(F_i|O)$ (the posteriors) using the output of a Bayesian feature detector.

$$P(F_i|O) = \frac{P(O|F_i)P(F_i)}{\sum_{F_j} P(O|F_j)P(F_j)} \quad (4)$$

### 1.2. Why Use Bayesian Feature Detectors?

Bayesian feature detectors can be combined to generate more flexible Bayesian feature detectors that are valid for large classes of images. Such feature detectors generate functions that take *apriori* probabilities and return *aposteriori* probabilities about some set of feature propositions. Arrays of these probabilities can act as generalized images in a larger system. A *generalized image* is an image-like array of measurements or scene parameters. Most image understanding systems use generalized images for intermediate stages in their calculations. E.g. an image may give rise to an array of edge elements to be linked into longer lines. [1]. In this paper I will also use the term generalized image for an image like array of probability distributions for scene parameters. That is because in my system such an array serves the same purpose as generalized images do in more conventional systems.

A Bayesian feature detector can be used to produce an array of probability distributions that can serve as a generalized image. Current operators often return a binary judgement about the presence of a feature at each point [2]. A generalized image whose elements are probability distributions contains more information than one whose elements are exact values.

## 2. Domains, Models, and Optimality

A *setting* of a generalized image is an assignment of truth or falsity to the set of feature propositions that the generalized image

245

consists of. A setting must be consistent with the mutual exclusivity of the feature propositions and their all encompassing properties. The systems I am describing consist of a particular set $G$ of generalized images that are relevant to the problem. A *circumstance* is a setting of each member of $G$. The universe consists of all possible circumstances. In the case I am interested in the universe is finite and each element of the universe corresponds to a state of the real world. Thus each circumstance has a probability in [0,1]. Thus the probability of a feature proposition being true given an observation is the sum of the probability of the circumstances where both the feature proposition and the observation's occurrence is true divided by the sum of all the probability of the circumstances where the observation occurs regardless of the feature proposition's truth. This is the number a Bayesian feature detector should determine. A truly optimal feature detector would take an observation and return this number.

Implicit in a feature detector are some assumptions about the structure of the image being observed. Formalizing this concept, I define a *domain* as a proposition from which a feature detector is constructed. To construct an example I define a set of *uniformly colored* rectangles as a set of rectangles whose interiors do not vary in color. Such a proposition is: "The scene is of overlapping uniformly colored rectangles viewed through additive Gaussian noise of standard deviation 4." Every domain has a model associated with it. A *model* consists of all circumstances that are consistent with the proposition of its domain.

A feature detector is a function that takes an observation and returns the probability of a feature proposition. An *observation* is a setting of certain generalized images. These generalized images are the *observed images*. I will only consider domains, $D$, whose propositions are powerful enough that for some mutually exclusive, all inclusive set of features, $F_i$, and any observation $O$ there is a well defined $P(O|F_i\&D)$ for all $F_i$. This $P(O|F_i\&D)$ is the sum of probabilities of the circumstances in $D$'s model where $O$ is observed and $F_i$ is true divided by the sum of the probabilities of the circumstances in $D$'s model where $O$ is observed. An *optimal Bayesian feature detector* is a function that takes an observation and calculates $P(O|F_i\&D)$. Equation (2) shows how to use an optimal Bayesian feature detector and a set of priors to calculate the probability of a feature proposition given an observation. A *Bayesian feature detector* returns probabilities that approximate the values that an optimal Bayesian feature detector would return.

### 2.1. Noise

For vision, domains usually contain probability statements. There are two sources of uncertainty that result in probabilistic descriptions.

(1) There are a variety of possible objects in the world that appear in unpredictable combinations.

(2) There are effects in the world that the domain does not account for since a domain is a simple description of a complex world.

The combination of these effects is what makes vision difficult. *Noise* is the second of these effects.

Noise in images, while unpredictable, has structure. Some of the structure can be determined from knowledge about the processes leading to noise. A large number of small additive noise sources will always result in Gaussian additive noise. Another way to determine noise is to take a large ensemble of images of known objects. From this one can determine what the probability of observing a certain color $o_i$ given that the a noiseless observation would yield color $a_j$. This results in an array of values $P(o_i|a_j)$ that I call the *structure of noise*. The *effect of noise* is difference between the observed image and how the image would have appeared if there had been no noise. Given knowledge about the structure that is being observed in an image it is possible to deduce facts about the effect of noise on that image.

### 3. A Rule for Combining Bayesian Feature Detectors

This section solves the problem of combining the Bayesian feature detectors based on two different domains $D_1$ and $D_2$ given that $D_1$ and $D_2$ are disjoint. Two domains are *disjoint* when the intersection of their models has zero probability. Two disjoint domains are :

(1) The scene is of overlapping uniformly colored rectangles viewed through additive Gaussian noise of standard deviation 4.
(2) The scene is of overlapping uniformly colored circles viewed through additive Gaussian noise of standard deviation 4.

These two domains' propositions can never be simultaneously true hence they are disjoint. Similar combination rules can be developed if the models are independent.

It is also necessary to have apriori knowledge about the probability of $D_1$ and $D_2$ being true in the image. All that is needed is the ratio of $\frac{P(D_1)}{P(D_2)}$. Then it is mathematically true that:

$$P\bigl[F\mid O\&(D_1+D_2)\bigr] = \frac{\begin{bmatrix} P(O\mid F\&D_1)P(D_1) \\ + \\ P(O\mid F\&D_2)P(D_2) \end{bmatrix}}{\begin{bmatrix} P(O\mid D_1)P(D_1) \\ + \\ P(O\mid D_2)P(D_2) \end{bmatrix}} \quad (5)$$

Thus it is possible to combine several Bayesian feature detectors to result in an feature detector that works on the union of their domains.

## 4. Generating Feature Detectors

I have developed two ways to produce feature detectors.

(1) Determine a function of the observed image and the priors that approximates the optimal Bayesian feature detector.
(2) Develop a Bayesian feature detector from an established operator.

### 4.1. A Simple Optimal Bayesian Feature Detector

I have written the program for an optimal Bayesian feature detector. There are two ways to view this feature detector.

(1) An optimal feature detector that detects the interior of uniformly colored regions against a random background.
(2) A Bayesian feature detector that attempts to detect boundaries but uses a constant for a number that the optimal Bayesian feature detector calculates at great computational cost.

Both the optimal interior detector and the boundary detector are based on these assumptions.

Sampling   Within a 3 by 3 window the 9 pixels can be considered as an ensemble of samples from a single point. Those samples are all the relevant information about the existence of a boundary at that point.

Monochromaticity   If the window lies in the interior of a region then all the pixels are of the same gray-level corrupted by noise.

Constancy of Noise   A single probability distribution describes the noise over the entire image. This probability distribution is known and additive.

Gray-levels   The distribution of gray-levels in a noiseless image would be uniform.

The optimal interior detector requires an assumption that results in a model with isolated uniformly colored regions in a sea of white noise.

Randomness   If the window lies outside a region then the values of the pixels are distributed according to the noise distribution. Every point lies within a uniformly colored region or in a random area.

Another way to view this feature detector is as a dector of boundaries between uniformly colored regions. Developing boundary detector requires an assumption about the behavior of points on the boundaries of regions.

Bimodality   Points (except for a negligible set) lie on the boundary of only two regions. The color of these

247

points is as the color of the interior points of one of these regions chosen at random with equal probability.

For a Bayesian feature detector it is necessary to calculate for a particular point $P(O \mid E(x,y))$ and $P(O \mid \text{not } E(x,y))$ where $O$ is the observed image and $E(x,y)$ is that $x,y$ is an exterior point. By the sampling assumption the probabilities need only be calculated from the 3 by 3 window containing the boundary point. By the randomness assumption $P(O \mid E(x,y))$ is just $\frac{1}{number\_colors^9}$. $P(O \mid \text{not} E(x,y))$ is

$$\sum_{c \in colors} \prod_{i,j \in window} P(O_{i,j} \text{ really is } c) \quad (6)$$

$P(O_{i,j} \text{ really is } c)$ is the probability that the observation at $i,j$ before it was corrupted by noise was color $c$.

The boundary detector requires that the probability that the distribution of observed gray-levels images are observed for a point on the boundary, $P(O \mid B(x,y))$ be determined, where $B(x,y)$ is the feature proposition that there is a boundary between two regions at $x,y$. This requires that the probability under all pairs of possible region colors needs to be considered. The distribution of possible observed colors according to bimodality assumption given two region colors is always flatter than the distribution of colors given the point lies on the interior of the region. I use the approximation of assuming the distribution is the same as the noise distribution. Thus I use the same number $\frac{1}{number\_colors^9}$ for $P(O \mid B(x,y))$. The sampling assumption determines that $P(O \mid \text{not} B(x,y))$ can be calculated from considering all possible region colors and for each multiplying the probability of the different pixels in the window observed. The probabilities of the different pixels of the window are calculated from table lookup on the noise distribution. I can use the same table for the entire image by the constancy of noise assumption.

### 4.2. Details for Generating the Probability of an Observation

The core of this operator is the calculation of $P(O_{i,j} \text{ really is } c)$. This calculation can be broken into two pieces:

(1) The probability that the gray-level that was observed would be observed if the actual gray-level at that point was $c$.

(2) The probability that the gray-level at that point was actually $c$.

The first piece is exactly the structure of the noise in the imaging process at that point, due to the constancy of noise assumption. The second piece can be deduced from the distribution of colors in the image, that is the image histogram. The image histogram is the result of the noise function being applied to the actual scene. It is the multiplication of the matrix $b_{i,j} = P(o_j \mid a_i)$ by the histogram of the actual colors (before noise) in the image. If one has the noise matrix then one can multiply the observed histogram by the inverse of this matrix to get what the actual distribution of colors might be.

### 4.3. The Implemented Operator

I have developed a program that implements the boundary detector described above. I have done work on implementing the technique that uses the image histogram to calculate the distribution of colors in the world. Figure 1 shows the results of applying different versions of the feature detector that assumes a uniform distribution of gray-levels to an aerial photograph.

## 5. Techniques for Approximating an Optimal Detector

### 5.1. Simplifying the Domain

*Simplifying the domain* is implementing a feature detector based on a more tractable domain than the one that describes the world knowledge of the situation. A more tractable domain allows one to develop more efficient optimal Bayesian feature detectors than the actual domain. The problem with using this technique is that it is undecidable how close the output of a feature detector based on the simplified model is to the output of a feature detector based on the actual model.

The randomness assumption is an example of such a simplification. In my original model, a boundary was a place where before noise the pixels were selected by

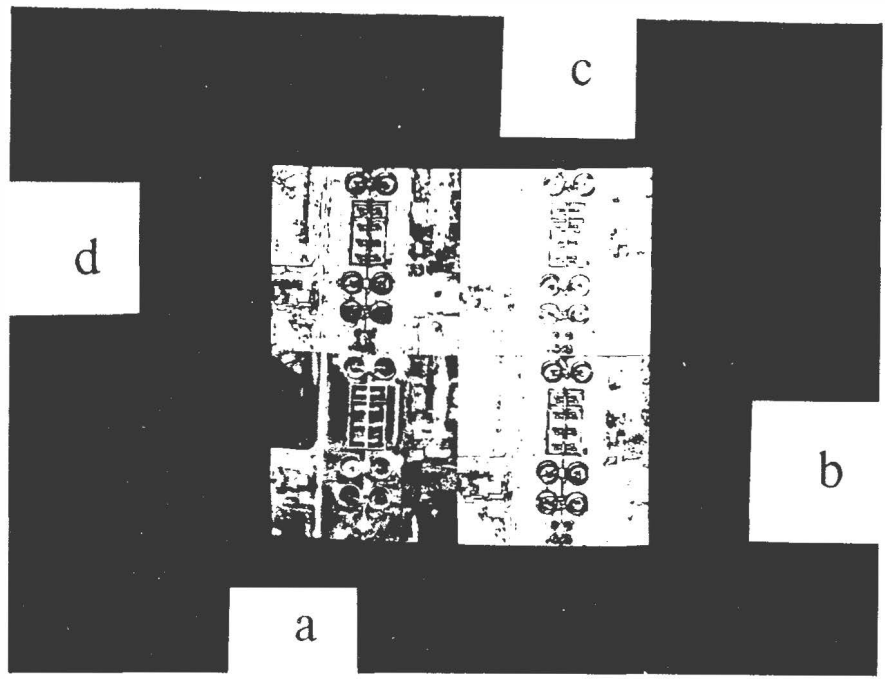

Figure 1
(a) is the aerial photo
(b) is output of boundary detector assuming Gaussian additive noise of $\sigma=8$
(c) is as (b) with $\sigma=16$
(d) is as (b) with $\sigma=4$

random selection between two gray-levels. The feature detector developed for the original model took about one million multiplications per pixel. The simplified model's feature detector takes about one thousand multiplications per pixel. However it no longer detects boundaries.

### 5.2. Limiting the Scope

*Limiting the scope* is using some small portion or easily calculated statistic is used to calculate the probabilities. Limiting the scope can be considered a easily analyzed form of simplifying the domain. One way of limiting the scope is to limit the range of observations relevant to the feature to a three by three window. This can also be considered a simplifying assumption about the independence of distant parts of the scene. Sampling out of the window without remembering structure treats the elements of the window as an ensemble. Such a feature detector only depends on a histogram of the window. The inaccuracy the the scope limiting assumption adds to the feature detector can be analyzed mathematically or experimentally.

### 5.3. Approximating the Function

*Approximating the function* is determining to what accuracy the probabilities need to be computed and only computing them up to that accuracy. Assume that the maximum error in the probability calculation that is allowable is .001. The *error* is the absolute difference between the output of the optimal detector and the output of the approximation. When calculating for the optimal boundary detector $P(O \mid notB(x,y))$ if for any $c$ in the summation there is an $i,j$ so that $P(O_{i,j} \text{ really is } c) < \frac{.001}{number\_colors}$ then it is unnecessary to calculate $\prod_{i,j \in window} P(O_{i,j} \text{ really is } c)$ since the sum of such cases is less than .001 .

### 6. Analysis of Established Operators

Some traditional edge detectors, line finders, circle detectors and the like output a number for every possible feature, and the

249

larger the number the stronger or more likely the feature. An feature detector is *monotonic in probability* under a set of assumptions if every threshold divides the cases below a certain probability from the cases above a certain probability.

To use such an operator as a Bayesian feature detector it is necessary to detect what priors and assumptions are necessary for the operator to be monotonic in probability. Then an approximation to the optimal operator can be constructed from the established operator and a table lookup or some simple function. The Bayesian feature detector developed is an approximation of the optimal Bayesian feature detector with the same scope.

### 6.1. Analysis of the Gradient

In this section the magnitude of the one dimensional gradient (absolute difference between two adjacent pixels) is analyzed as a feature detector. The one dimensional gradient is the difference between two adjacent pixels. If a boundary lies between two adjacent pixels then they come from different regions. Thus the gradient can measure the probability that the two pixels come from different regions. I shall demonstrate a simple set of models for which the gradient is monotonic in probability. Such models assume uniformly colored regions with constant noise as the optimal detector in the previous section did. They also assume that the distribution of colors in the observed image is the constant distribution and that the prior probabilities for the boundary at each point is a constant. These assumptions simplify the problem to a great extent. Since only two points are used by the one dimensional gradient the probability of the existence of a boundary can be calculated by the following technique. Let the probability that the actual gray-levels (uncorrupted by noise) of the two points are from two randomly selected uniformly colored regions be $P(W|B)$. Let the probability that the actual color of the two points is from the same uniformly colored region be $P(W|NB)$. Let the prior probability of the existence of a boundary be $P(B)$. Then the formula for calculating the probability of the two points coming from different regions is:

$$\frac{P(W|B)P(B)}{P(W|B)P(B)+P(W|NB)(1-P(B))}$$

Let $o_1$ and $o_2$ be the two pixels observed by the gradient. Then the gradient magnitude is $|o_1-o_2|$. $P(W|NB)$ can be calculated by $\sum_{c \in colors} P(o_1|c)P(o_2|c)$ where $P(o_i|c)$ is the probability that the observed color will be $o_i$ given that the actual color is $c$. $P(W|B)$ is

$$\sum_{c_1 \in colors, c_2 \in colors} P(o_1|c_1)P(o_2|c_2)$$

which is

$$\left[\sum_{c \in colors} P(o_1|c)\right]\left[\sum_{c \in colors} P(o_2|c)\right]$$

Thus under these assumptions the probability is determined by the distribution of $P(o|c)$ which is the known structure of the noise and the prior $P(B)$.

### 6.2. Constraints on the Noise Distribution

$f$ is *monotonic* with $g$ when:

$$f(X)=f(Y) \rightarrow g(X)=g(Y)$$
$$f(X)>f(Y) \rightarrow g(X)>g(Y)$$
$$f(X)<f(Y) \rightarrow g(X)<g(Y)$$

Monotonicity is an equivalence relation by this definition. For any functions $P(W|B)$ and $P(W|NB)$ the resulting set of functions that calculate the probability of boundaries from windows for all $P(B)$ will be monotonic in each other. Thus monotonicity of the gradient in an optimal boundary detector in a two pixel window is dependent only on the structure of the noise.

If the optimal operator on two points is monotonic with the gradient for some value of $P(B)$, it is monotonic for every value of $P(B)$. I can assume $P(B)$ is a constant over the entire generalized image. Thus the gradient has to be monotonic with this function:

$$1 - \frac{\sum_{c \in colors} P(o_1|c)P(o_2|c)}{\left[\begin{array}{c}\sum_{c \in colors} P(o_1|c)P(o_2|c) \\ + \\ \sum_{c \in colors} P(o_1|c) \sum_{c \in colors} P(o_2|c)\end{array}\right]}$$

For $o'' > o' \geq o$ these equations reduce to:

$$\frac{\sum_{c \in colors} P(o''|c)P(o|c)}{\sum_{c \in colors} P(o''|c)} < \frac{\sum_{c \in colors} P(o'|c)P(o|c)}{\sum_{c \in colors} P(o'|c)}$$

The numerators of these equations are $P(W|NB)$ and the denominators are the probability of observing the color in the observed image. This is a necessary and sufficient condition for monotonicity with the gradient.

### 6.3. Simplifying Assumptions about the Noise Distribution

Constraining the possible noise functions in various ways can further simplify the problem. There are two sets of constraints that I find interesting because they show important things about when the gradient will succeed.

(1) $\forall o \sum_{c \in colors} P(o|c) = C$ where $C$ is a constant and $P(o|c)$ is symmetric in $o$ about $c$. The noise is additive. *Additive noise* is when $P(o|c)$ is a function of $o-c$.

(2) The noise is the weighted sum of two noise functions, one of which is additive to the data the other replaces the data. The ratios of the standard deviations and the weights in the sum are used to build the noise distribution.

If the noise is additive and unimodal then the gradient is unimodal in the optimal operator in two points. The most popular assumption about noise is that it is Gaussian and additive. This kind of noise fits these assumptions, thus for Gaussian additive noise the gradient is monotonic in the optimal Bayesian feature detector in two points.

A combination of replacement noise and additive noise is consistent with the gradient only when the result of the combination is additive noise. There is reason to believe that the gradient will only be monotonic in the optimal two point Bayesian feature detector when the noise is additive.

Any convolution based operator is similar to the gradient in this respect. Thus a convolution based operator will be monotonic in probability only when the noise is additive.

### 6.4. Uses of Models of Established Operators

Because they only depend on the magnitude of the gradient $P(W|B)$ and $P(W|NB)$ can be calculated by table lookup on the magnitude of the gradient. This means that there is a way to measure the local applicability of the gradient on the image. This can be useful because in those cases where the gradient is inapplicable some more sophisticated operator can be called upon. Also this study allows one to identify the situations when the gradient as a boundary detector works. This means that when operators are being developed for a new domain these features of the domain need only be checked. However if the noise is not additive or unimodal then the gradient may have some difficulty as a boundary detector. Similar analyses can be done for operators other than the gradient to determine more precisely their domains of application.

### 7. Conclusion

In this paper Bayesian feature detectors are described. They are useful because they can be developed more naturally from the definitions of the features and their accuracy can be proven. Bayesian feature detectors' output is easy to interpret and the output of several such detectors can be easily combined. An optimal detector for a simplified model has been implemented. The conditions under which the one dimensional gradient is effective are analyzed by comparing its output to that of a Bayesian feature detector. The gradient was theoretically found to be effective only if the noise was additive (experimentation is pending). This result can be extended to other convolution based operators. It was predicted on theoretical grounds that the gradient is effective when the noise is symmetric and unimodal. Experiments are planned.

### References

1. D. H. Ballard and C. M. Brown. in *Computer Vision*. Prentice-Hall Inc.. Englewood Cliffs. New Jersey. 1982. 14,63,85-86.

2. J. F. Canny. Finding Edges and Lines in Images. 720. MIT Artificial Intelligence Laboratory. June 1983.

This work was supported in part by the Defense Advanced Research Projects Agency under grant number N00014-82-K-0 193 and in part by the National Science Foundation Graduate Fellowship grant number RCD-8450125.